\documentclass[pdflatex,sn-mathphys-num]{sn-jnl}

\usepackage{graphicx}%
\usepackage{multirow}%
\usepackage{amsmath,amssymb,amsfonts}%
\usepackage{amsthm}%
\usepackage{mathrsfs}%
\usepackage[title]{appendix}%
\usepackage{xcolor}%
\usepackage{textcomp}%
\usepackage{manyfoot}%
\usepackage{booktabs}%
\usepackage{algorithm}%
\usepackage{algorithmicx}%
\usepackage{algpseudocode}%
\usepackage{listings}%

\theoremstyle{thmstyleone}%

\theoremstyle{thmstyletwo}%

\theoremstyle{thmstylethree}%

\begin{document}

\title[Article Title]{EEG-Based Consumer Behavior Prediction: An Exploration from Classical Machine Learning to Graph Neural Networks}

\author[1]{\fnm{Mohammad Parsa} \sur{Afshar}}\email{m.parsa.afshar@ut.ac.ir}

\author*[1]{\fnm{Aryan} \sur{Azimi}}\email{aryanazimi@ut.ac.ir}

\affil*[1]{\orgdiv{College of Engineering}, \orgname{University of Tehran}, \orgaddress{\city{Tehran}, \country{Iran}}}

\abstract{Prediction of consumer behavior is one of the important purposes in marketing, cognitive neuroscience, and human-computer interaction. The electroencephalography (EEG) data can help analyze the decision process by providing detailed information about the brain's neural activity. In this research, a comparative approach is utilized for predicting consumer behavior by EEG data. In the first step, the features of the EEG data from the NeuMa dataset were extracted and cleaned. For the Graph Neural Network (GNN) models, the brain connectivity features were created. Different machine learning models, such as classical models and Graph Neural Networks, are used and compared. The GNN models with different architectures are implemented to have a comprehensive comparison; furthermore, a wide range of classical models, such as ensemble models, are applied, which can be very helpful to show the difference and performance of each model on the dataset. Although the results did not show a significant difference overall, the GNN models generally performed better in some basic criteria where classical models were not satisfactory. This study not only shows that combining EEG signal analysis and machine learning models can provide an approach to deeper understanding of consumer behavior, but also provides a comprehensive comparison between the machine learning models that have been widely used in previous studies in the EEG-based neuromarketing such as Support Vector Machine (SVM), and the models which are not used or rarely used in the field, like Graph Neural Networks.}

\keywords{Neuromarketing, Machine Learning, Deep Learning, Graph Neural Network, Classical Algorithms, Electroencephalography}

\maketitle

\section{Introduction}\label{sec1}

Understanding and predicting consumer behaviour is one of the key issues in today’s world for businesses and companies [1], due to its importance in product design [2], advertisements, and user experience [3]. Using brain data, machine learning, and data mining methods would lead to data-driven decisions. This can be more helpful for a better decision-making [4] process than using the classical methods. The classical data gathering methods, such as surveys, field studies, and interviews [5], [6], [7] in spite of being applied, have biases and limitations [8], [9], [10] like outcome bias and halo effect [11]. In this condition, using neuro signals can provide a clearer picture of the decision-making process [12].

As purchasing decisions and consumer behaviour play a crucial role and have a significant impact on the success or failure of a business, a deeper understanding of this process has become particularly important for marketing [13], [14], [15], [16] and product development [17], [18]. It is necessary for companies to gain knowledge about their consumers behavior to optimize their marketing strategies [19] and develop and better user experience [20] for costumers, especially in today’s competitive markets [21]. Furthermore, by gaining insight of consumers opinion about new products or marketing campaigns, companies can make competitive advantages and reduce their costs [22].

Many different tools and method have been utilized to measure brain activity or brain response to a stimulus in tasks related consumer neuroscience. These methods include Functional magnetic resonance imaging (fMRI) [23], [24], Functional near-infrared spectroscopy (fNIRS) [25], Magnetoencephalography (MEG) [26], and Electroencephalography. Each of these methods has its own advantages and disadvantages. For example, fMRI has high spatial resolution [27] but it is expensive [28] and limited to laboratory environments, while EEG, due to its portability [29], lower cost [30], and high time resolution [5], [9], [31], [32], has become one of the most useful tools for measuring the brain activity related to decision-making in consumer neuroscience [33], [34], [35].

EEG can record the rapid changes in brain activity by measuring the electrical potential difference caused by the activity of neurons [29]. Due to this feature, it is possible to measure the brain responses to marketing stimuli and products, and detect brain patterns that are related to buying preferences [9]. Because of this reason, EEG plays an important role in developing consumer behavior prediction and, the combination with machine learning methods, can provides a deeper insight into the mechanisms which are related to the cognitive process [36].

Analyzing EEG signals in this field has usually been done by classical machine learning algorithms such as Support Vector Machine (SVM), Random Forest (RF), and k-Nearest Neighbors (KNN), and deep learning models such as Convolutional Neural Network (CNN) and Long Short-Term Memory (LSTM) [37]. These models usually use features from the time, frequency, and time-frequency domains, and they do are not focusing on modelling the interactions between brain channels, while these interactions can be shown as a graph in which node are electrode and edges are the interactions. The graph-based modelling can be helpful for a better understanding of the decision-making process [38], [39].

Graph Neural Network (GNN) based models are specifically designed for graph-structured data and allow learning topology-based representations [40]. In the context of EEG, it is possible to assign nodes to electrodes and edges to connectivity values and use GNNs for extracting cognitive-related patterns [38], [41], [42]. However, GNNs have been used in different EEG tasks, such as emotion detection [43], [44], [45], seizure detection [46] and Alzheimer detection [47]; they have not been widely used in EEG-based neuromarketing tasks.

One of the major gaps in neuromarketing research is the lack of application of graph-based models like graph neural networks. Most previous studies have used classical machine learning methods or neural network models such as CNN and LSTM. Although these models are successful in extracting time-frequency features, they are unable to explicitly model the connectivity structure between brain channels, even though these interactions and connections play a key role in cognitive tasks and consumer decision-making [48]. Furthermore, another limitation in previous studies in the field is the lack of comprehensive comparisons evaluating a diverse range of models and comparing them.

This research proposed a framework for consumer behavior prediction by using EEG signals, 
with the aim of addressing these gaps and introducing several key contributions as follows:

\begin{itemize}
    \item Modelling brain connectivity, transforming EEG data to graphs, and extracting connectivity features for GNNs.
    \item Utilizing a diverse range of GNNs architectures and layers, such as Graph Convolutional Network (GCN), Graph Attention Networks (GAT), and Graph Sample and Aggregate (GraphSAGE).
    \item Using Classical models, which have not been widely implemented in other EEG-based neuromarketing studies, such as XGBoost and LightGBM.
    \item Designing and implementing an ensemble stacking model combining different classical algorithms.
    \item Comparing the performance of classical models trained on common EEG extracted features, as well as evaluating and comparing different GNN architectures based on brain connectivity features in consumer behavior classification.
\end{itemize}

To increase reliability and repeatability, the NeuMa dataset [49] is used in this research, which is publicly available. Results have shown that there is no significant difference between the models, but overall, GNNs achieve better performance on the minority and unbalanced classes.

The research is structured as follows: In the Related Works section, previous studies and gaps will be explained. In the Methodology, the dataset, EEG preprocessing, and models will be introduced. The training settings and models' architectures will be explained in the Experiments section. In the Results section, the results of evaluating models with different metrics will be discussed, and a comparison between the models will be provided. The Discussion section will provide theoretical and marketing implications, and Conclusions section will focus on the findings, limitations, and future directions of this study.

\section{Related works}\label{sec2}

In recent decades, neuromarketing research has increasingly utilized novel methods to understand the consumer decision-making process. EEG is recognized as a key tool in the field, due to its high temporal resolution, cost-effectiveness, and lower complexity compared to imaging methods. Previous studies tried to use time domain, frequency domain, or time-frequency domain features of EEG signals to extract indicators related to attention, motivation, or product evaluation and correlate them with behavioral variables. A summary of the previous key studies is provided in Table ~\ref{tab:eeg_summary}.

\subsection{Datasets, stimuli, objectives, and task types}\label{subsec2}

EEG-based neuromarketing studies have been used with different datasets, stimuli, objectives, and task types. Stimuli include a range of visual [50], [51], [52], [53], [54], taste [55], [56], olfactory [57], [58], [59], and auditory stimuli [60], [61]. Visual stimuli are the most used [48], and they include promotional and product images and videos, as well as product pages in a web environment simulation. Taste stimuli contain edible products; olfactory stimuli include gas sampling and food samples [57], and musical pieces are used as auditory stimuli [60]. In addition, audiovisual stimuli are also used. 
Previous studies in this field have been defined by different objectives, such as mobile phone [51], [62], sports shoes [52], coffee brand preferences [55], and user behavior in web stores [53]. These objectives are associated with different task types, including classification of binary buy or not buy decisions [50], preference and like/dislike ratings [52], identifying willingness to pay [63], and predicting like and view rates [64]. Furthermore, other data, such as eye tracking and physiological data [65], have been used as supplementary data for EEG signals.

\subsection{Extracted features}\label{subsec2}

The dominant approach in feature extraction is based on three domains: time, frequency, and time-frequency [54], [66]. For the time domain, Hjorth Parameters [52], [58], [59] and Riemannian Geometry [67] are the common methods, while Fast Fourier Transform (FFT) [68], [69] and Power Spectral Density (PSD) [69], [70], [71], [72] are used as frequency domain features. Moreover, for the time-frequency domain, the Discrete Wavelet Transform (DWT) [73], [74], [75], [76] is used. 

In recent years, functional connectivity analysis has been widely used as a network-level feature in EEG-based tasks. Pearson correlation, spectral coherence, fuzzy measures such as PLV, PLI and wPLI, and directional indices such as DTF and PDC are used to model information flow between electrodes [77]. By using these methods, a graph can be constructed that has information about brain activity in response to stimuli.

\subsection{Models: Classical, Ensemble, and Neural Networks}\label{subsec2}

Classical models like Support Vector Machine (SVM) [51], [54], [66], [68], [74], [78], Random Forest (RF) [72], [74], [76], [79], and k-nearest neighbors (KNN) [52], [65], [74], [78] are the most common models in neuromarketing. The key advantage of these models is low computational cost; also, they do not require huge datasets compared with neural network models. Various ensemble models are also used in the EEG analysis tasks with different architectures. Furthermore, principal component analysis (PCA) [53], [61], [75] is commonly used for the dimensionality reduction of EEG data.

In the neural network models, convolutional neural network (CNN) [55], [75], Multilayer perceptron (MLP) [64], long short-term memory (LSTM) [71], [76], [80], and deep architectures [72], [79], [81] are utilized in the EEG-based neuromarketing tasks, while recurrent neural network (RNN) [82], CNN-RNN [83], and graph neural network (GNN) have been used in other EEG-based neuroscience tasks. GNN models can be applied to benefit from the brain connectivity network. Graph Convolutional Network (GCN) [44], [45], Graph Attention Networks (GAT) [84], and GraphSAGE [85] are the common models used in the previous studies for EEG analysis.

\begin{table}[h]
\caption{Summary of EEG-based consumer behavior prediction studies}\label{tab:eeg_summary}
\centering
\renewcommand{\arraystretch}{1.5}
\begin{tabular}{@{}p{0.5cm} p{0.5cm} p{2.5cm} p{3cm} p{5cm}@{}}
\toprule
\textbf{Ref. No.} & \textbf{Year} & \textbf{Model} & \textbf{Dataset} & \textbf{Key Findings} \\
\midrule
{[51]} & 2021 & Artificial Neural Network (ANN), SVM & EEG signals from 11 subjects, viewing 20 smartphone images & The combination of PCA, Artificial Bee Colony, and ANN achieved the highest accuracy with an average 72\%. \\
{[52]} & 2022 & KNN, SVM & EEG dataset collected from 15 subjects, viewing 25 sports shoes & Differential Entropy features with KNN achieved the highest accuracy (94.22\%). \\
{[54]} & 2022 & SVM, KNN, Naive Bayes (NB), Decision Tree (DT) & EEG signals from 20 subjects, viewing 3 types of advertisements & SVM achieved the highest accuracy with 87\% for predicting affective attitude and 84\% for predicting purchase intention. \\
{[63]} & 2023 & Deep Learning Network (spectral, spatial, temporal layers) & EEG data from 213 subjects, viewing 72 product images & Deep learning model achieved about 85\% prediction accuracy. \\
{[71]} & 2023 & KNN, SVM, Bidirectional-LSTM & EEG data from 25 subjects, viewing 42 product images & Bidirectional-LSTM achieved the highest accuracy (96.83\%) using frontal region signals. \\
{[75]} & 2024 & CNN–LSTM, SVM, RF, CNN-Transformer & EEG dataset collected from 25 subjects, viewing 42 product images & CNN–LSTM model achieved the highest accuracy (97.12\%). \\
{[59]} & 2024 & Gradient Boost (GB), NB, KNN, DT, SVM, RF, CNN & EEG data from 33 participants, exposed to scented and unscented product boxes & SVM with Lempel–Ziv Complexity feature achieved 93\% accuracy, while CNN achieved 91.6\% accuracy. \\
{[53]} & 2024 & SVM, RF, KNN, Shallow Neural Network & EEG data from 66 participants, for 5 products across 6 categories & SVM achieved the highest accuracy (87.1\%) in predicting purchasing decisions. \\
\botrule
\end{tabular}
\end{table}

\subsection{Research gaps and contributions of this research}\label{subsec2}

Despite significant advances in the EEG-based neuromarketing studies, the existing literature has several structural limitations, including restricted use of machine learning models, dimensionality reduction and feature selection methods, and the lack of a comprehensive comparison between different models. In this research, in addition to utilizing a wide range of classical models such as XGBoost and LightGBM, which are used in other tasks related to EEG signals in neuroscience fields [42], [82], different dimensionality reduction and feature selection methods are utilized. Further, an ensemble stacking model is designed and implemented to take advantage of combining multiple algorithms. In this research, brain connectivity features were extracted and represented as data in a graph structure. This graph representation allows the use of GNNs, and varied GNN architectures, including GCN, GAT, and GraphSAGE, have been utilized.

This diversity in methods and models allows for a comprehensive comparison between the performance of algorithms. This comparison is carried out in this study. Such an approach not only covers the weaknesses of previous research but also opens up new horizons in EEG data analysis in neuromarketing.

\section{Methodology}\label{sec3}
\begin{figure}[h]
    \centering
    \includegraphics[width=0.8\textwidth]{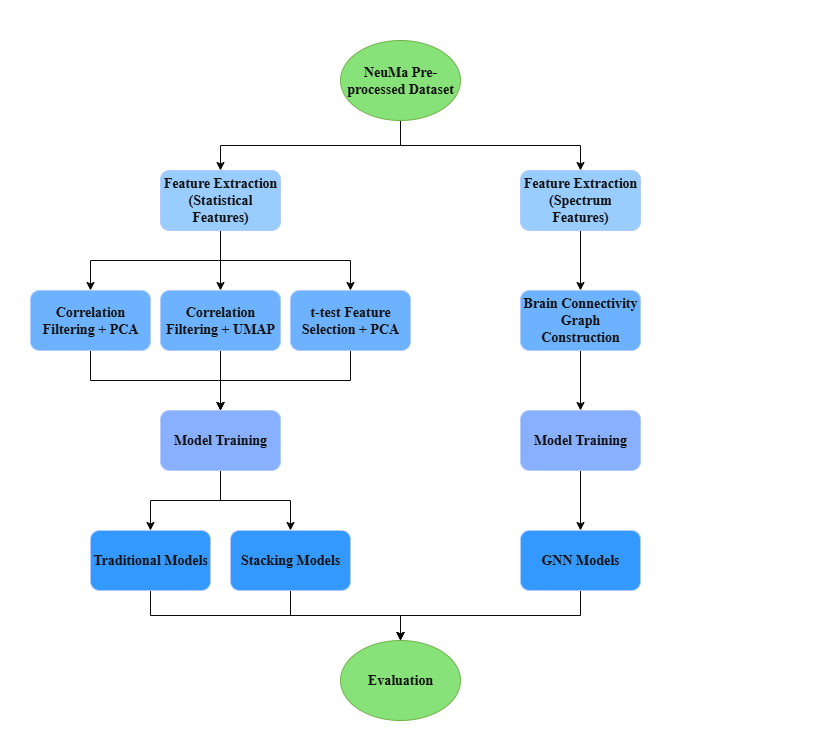}
    \caption{Overview of the proposed methodology for EEG-based consumer behavior prediction}
    \label{fig:methodology}
\end{figure}

The methodology of this research is illustrated in Figure~\ref{fig:methodology}. In the next subsections, the dataset, different pipelines, various machine learning models, and deep learning models will be explained.

\subsection{Dataset}\label{subsec2}

For this study, the pre-processed version of the NeuMa dataset [49], which is publicly available, was used. The NeuMa dataset is a multimodal EEG-based neuromarketing dataset designed to investigate consumer decision-making behavior. This dataset contains data from 42 healthy adult participants (23 males and 19 females, age 31.5 ± 8.84 years), all native Greek speakers, engaged in a supermarket brochure browsing task (six pages, 24 products per page, totaling 144 products). Participants could select the products they intended to purchase while their brain activity (EEG) and eye movement were recorded, along with behavioral and questionnaire data.

The products are labelled in two different classes, “Buy” and “NoBuy”. Furthermore, on average, each participant selected 18 products, and 126 products remained unselected. The EEG associated with each product was separated during the period when the participant viewed it. EEG signals were pre-processed with a Butterworth bandpass filter (0.5 to 45 Hz, 3rd order, zero phase) and artifact removal using Artifact Subspace Reconstruction (ASR) [86] and FORCe [87].

The brain activity was recorded using 21 dry EEG sensors according to the international 10–20 system, including two reference electrodes, with a sampling rate of 300 Hz. In the pre-processed NeuMa dataset, 19 channels covering the main EEG electrodes were provided for analysis, while the reference channels were removed.

\subsection{Classical Models}\label{subsec2}

\subsubsection{Feature extraction}\label{subsubsec2}

Preparing the EEG data for classical machine learning has several steps, which include filtering, segmentation, and feature extraction. In the first step, EEG signals were filtered using a third-order Butterworth bandpass filter. This filtering was used for each of the standard frequency bands, including delta, theta, alpha, beta, and gamma, and the distinct frequency components of each EEG channel were extracted.

In the next step, the data were segmented according to the structure of the study. Each sample is related to a user, a page, and a product. Only the segments longer than 0.5 seconds were considered in the analysis to ensure feature stability and reliability. For each valid segment, spectral features were calculated from two perspectives: (1) Fast Fourier Transform (FFT) to compute the frequency domain and extract the statistical indicators of mean, standard deviation, skewness, and kurtosis, and (2) Power Spectral Density (PSD) using the Welch method with the same statistical indices extracted. Finally, for each EEG channel and frequency band, eight statistical features were computed.

\subsubsection{Dimensionality Reduction and Data Preparation Pipelines}\label{subsubsec2}

Since the final data is a large matrix with 760 features, there was a need to reduce dimensionality. This process had been done in three different pipelines: (1) Using correlation between features and principal component analysis (PCA), a correlation coefficient greater than 0.90 between pairs of features was used, then PCA was applied in a way that more than 90\% of the data variance was retained; (2) Using correlation method and uniform manifold approximation and projection (UMAP) for representing data in 50-dimensional space; (3) the top 100 features were selected based on an independent t-test (with a significance threshold of 0.05) and then PCA was applied to this set, which explained more than 95\% of the variance.

\subsubsection{Classical Models Descriptions}\label{subsubsec2}

To classify EEG data into two different classes, “Buy” and “NoBuy,” a wide range of classical machine learning algorithms was employed. These models included logistic regression (LR), k-nearest neighbors (KNN), support vector machines (SVM) with RBF kernel, random forests (RF), gradient-based algorithms including XGBoost and LightGBM, and two probabilistic models, Naïve Bayes (NB) and Gaussian Process Classifier.

The diversity of these models makes it possible to analyze the performance of different algorithmic categories (linear, tree-based, distance-based, and probabilistic). The models were trained using different pipelines with different dimensionality reduction and feature selection methods to provide an analysis of the models’ performance in each pipeline.

\subsubsection{Stacking Ensemble Models}\label{subsubsec2}

To benefit from the strengths of different classical models, a stacking ensemble model was utilized to combine the models' predictions. Four different models were trained, and the final predictions were made by a meta model. Each of the four base model make an initial prediction, and the meta model predicts by feeding from the base models' predictions. Diverse base models have been used, including LR, KNN, NB, and LightGBM, which increases the variety of prediction methods. Furthermore, XGBoost has been utilized as the meta model to provide the final predictions.

By using this architecture for the stacking model, it can be able to model complex non-linear patterns, as well as the linear interactions. These stacking models can provide a deeper insight into combining different models to improve performance.

\subsection{Graph Neural Network Models}\label{subsec2}

\subsubsection{Feature Extraction}\label{subsubsec2}

In contrast with the classical model features, the spectral features of EEG signals were extracted and used for training the GNNs. These include the Fourier domain (FFT) and the power spectral density (PSD) features from the delta, theta, alpha, beta, and gamma bands. Similar to the classical models' feature extraction method, a third-order Butterworth bandpass filter and a channel segmentation method were implemented. 

\subsubsection{Brain Connectivity Modelling}\label{subsubsec2}

For modelling the brain connectivity and constructing graphs, the EEG electrodes were considered as graph nodes, and the Pearson correlation coefficient was used to calculate the edge features. This coefficient is defined as follows:

\begin{equation}
r_{ij} = \frac{\sum_{k=1}^{n} \left(x_{ik}-\bar{x_i}\right)\left(x_{jk}-\bar{x_j}\right)}
{\sqrt{\left(\sum_{k=1}^{n} \left(x_{ik}-\bar{x_i}\right)^2\right)
\left(\sum_{k=1}^{n} \left(x_{jk}-\bar{x_j}\right)^2\right)}}
\end{equation}

where $x_{ik}$ is the $k$-th feature of electrode $i$, $\bar{x_i}$ is the mean of the features of electrode $i$, and $n$ is the total number of features.

As a result of this, a correlation matrix has been constructed for each data set, which represents the connectivity between each pair of nodes. This can help consider both spectral features and the interaction between brain regions.

\subsubsection{GNN Architectures and Model Descriptions}\label{subsubsec2}

A wide range of graph neural network architectures were implemented to analyze the graphs which have been constructed using EEG-brain features and Pearson correlation coefficient. These models contain baseline models and layers such as Graph Convolutional Network (GCN), Graph Attention Networks (GAT) and GraphSAGE. While GCN can represent new features for each node by using the local connections and nearby nodes, GAT uses an attention mechanism to put more weight on important edges which can define the importance of each brain region. Also, GraphSAGE uses a method that have a significant impact on graphs with high dimension node features.

In addition to baseline models, more advanced architectures were also utilized for classifying the consumer decision. A Residual architecture with the skip connections mechanism was used to reduce the information loss in deeper layers. Also, multilayer models were introduced, which combine different types of layers to take advantage of different architectures and their combination. Furthermore, lighter architectures were used to reduce computational complexity, and deeper architectures were utilized to enhance the representational capacity of the models.

To train and evaluate the models, K-fold cross-validation was used; each model was trained on the data of each fold and then predictions were made on the test data. Finally, the prediction results of each fold were merged together in the form of majority voting, and the major votes obtained was considered as the final results of the model.

\section{Experiments}\label{sec4}

This section aims to provide details of different classical, ensemble, and graph neural network architectures. In these experiments, several different pipelines are defined, each based on a different combination of feature extraction, dimension selection or reduction, and modelling steps. In the following, details regarding the implementation, experimental setup, and features used in each class of models are provided.

\subsection{Setup and Implementation Details}\label{subsec2}

All implementations were done in Python 3.10.18, and standard machine learning and deep learning libraries, including scikit-learn, XGBoost, LightGBM, PyTorch, and PyTorch Geometric, were used for development and execution. The classical models were run on the CPU (AMD Ryzen 77435HS), while the deeper, graph-based models were trained on the GPU (NVIDIA RTX 3050, 4GB VRAM).

\subsection{Classical and Ensemble Models}\label{subsec2}

\subsubsection{Feature Extraction Pipelines}\label{subsubsec2}

Three different pipelines were designed for data preprocessing and dimensionality reduction:

\begin{itemize}
    \item \textbf{Pipeline A:} Remove high correlations (threshold $|r| > 0.9$), normalize with Standard Scaler, and perform dimensionality reduction with PCA based on 90\% of explained variance.
    \item \textbf{Pipeline B:} Remove high correlations, normalize with Standard Scaler, and perform dimensionality reduction with UMAP to 50 components.
    \item \textbf{Pipeline C:} Select top 100 features with an independent t-test ($p < 0.05$), normalize with Standard Scaler, and perform dimensionality reduction with PCA based on 95\% of explained variance.
\end{itemize}

These three pipelines allowed the model evaluation under different dimensionality reduction conditions.

\subsubsection{Hyperparameter Selection and Optimization}\label{subsubsec2}

For all models, Grid Search was used with 5-fold cross-validation. The main optimization criterion was the weighted F1 to better handle the class imbalance issue. Other criteria were also used to report the final results.

To prevent data leakage, all normalization, dimensionality reduction, and feature selection steps were performed only on the training data and then applied to the test data.

\subsubsection{Classical Models}\label{subsubsec2}

A set of classical machine learning methods, including logistic regression, K-nearest neighbors, random forest, XGBoost, LightGBM, support vector machine with RBF kernel, Naive Bayes, and Gaussian process classifier, was used. These models have been previously used in areas related to EEG signal analysis, but have been less systematically evaluated in the context of neuromarketing.

\subsubsection{Ensemble Models}\label{subsubsec2}

To combine the power of different models, a stacking framework was designed, which includes:

\begin{itemize}
    \item Base models: Naive Bayes, Logistic Regression, KNN, LightGBM
    \item Meta-learner: XGBoost
\end{itemize}

The use of 5-part StratifiedKFold cross-validation ensured that out-of-fold predictions were generated correctly for training the meta-model.

\subsection{Graph Neural Network Models}\label{subsec2}

\subsubsection{Feature representation and graph construction}\label{subsubsec2}

For each EEG sample, spectral features were extracted in five frequency bands, resulting in 98-dimensional vectors for each electrode. The electrode placement system followed the international standard 10–20, with 19 nodes (electrodes) per graph. To model brain connectivity, a fully connected graph was used, in which edge weights were calculated as the Pearson correlation coefficient between the spectral features of the electrodes

\subsubsection{Graph Neural Network Architectures}\label{subsubsec2}

This study evaluated a diverse set of GNN architectures, ranging from basic and lightweight models to deep and hybrid architectures. The purpose of this diversity was to investigate the ability of each architecture to extract spectral features and inter-electrode connections. The models studied include the following:

\begin{itemize}
    \item \textbf{BaselineGCN} – consisting of two GCN layers with ReLU and Dropout activators, along with a linear layer for classification.
    \item \textbf{BaselineGAT} – consisting of two GAT layers with a multi-head attention mechanism (4 heads in the first layer and 1 head in the second layer), an ELU activator, and a final linear layer.
    \item \textbf{BaselineSAGE} – consisting of two GraphSAGE layers for neighborhood information aggregation, along with Dropout and ReLU activators.
    \item \textbf{ResidualGCN} – an extended version of BaselineGCN with residual connections to improve gradient flow in deeper models.
    \item \textbf{HybridModel} – a combination of MLP, GCN, and GAT layers in sequence, to simultaneously cover local features and graph dependencies.
    \item \textbf{RegularizedGNN} – alternating architecture of GCN and GAT with LayerNorm and structured classifier, to improve generalizability.
    \item \textbf{LightweightGCN} – very simple architecture with only one GCN layer and one linear layer, as a lightweight baseline.
    \item \textbf{BalancedGAT} – a compact model with one input MLP layer, one GAT layer, and one final linear layer, suitable for lightweight attention-based modelling.
    \item \textbf{MultiGNN} – multi-branch architecture consisting of three parallel paths (GCN, GAT, and GraphSAGE) with different Pooling methods, which are finally combined by a classification layer.
    \item \textbf{ResidualAttentionGNN} – consisting of two GAT layers with residual connections and a learnable attention mechanism for weighting features.
    \item \textbf{DeepGNN} – deep hierarchical architecture with three blocks including parallel GCN, GAT, and GraphSAGE sub-sections, along with residual connections and a multilayer classifier.
\end{itemize}

\subsubsection{Training Settings}\label{subsubsec2}

All models were trained using the AdamW optimizer (learning rate = 0.001, weight decay = 0.01, batch size = 32). For adaptive adjustment of the learning rate, ReduceLROnPlateau was used, which reduces the learning rate by a factor of 0.5 if the validation accuracy does not improve within 5 cycles. The models were trained for a maximum of 100 cycles, with an early stop at a threshold of 15 cycles (based on validation error). The loss function used was class-weighted cross-entropy. Summary of each model provided in Table~\ref{tab:gnn_arch}.

\begin{table}[h]
\caption{Description of the GNN architectures used in this study}\label{tab:gnn_arch}
\centering
\renewcommand{\arraystretch}{1.5}
\begin{tabular}{@{}p{3cm} p{3cm} p{2.5cm} p{3.2cm}@{}}
\toprule
\textbf{Model} & \textbf{Layers} & \textbf{Key Features} & \textbf{Special Mechanisms} \\
\midrule
BaselineGCN & 2 GCN + 1 Linear & Simple spectral propagation & ReLU, Dropout \\
BaselineGAT & 2 GAT + 1 Linear & Multi-head attention & 4+1 heads, ELU \\
BaselineSAGE & 2 SAGE + 1 Linear & Neighborhood sampling & ReLU, Dropout \\
ResidualGCN & 2 GCN + Residual + 1 Linear & Improved gradient flow & Skip connection \\
HybridModel & MLP + GCN + GAT + GCN + Linear & Sequential hybridization & BatchNorm, Dropout \\
RegularizedGNN & 1 GCN + 1 GAT + 2 Linear & Normalization & LayerNorm, structured classifier \\
LightweightGCN & 1 GCN + 1 Linear & Minimal baseline & Fast, low-complexity \\
BalancedGAT & MLP + GAT + Linear & Compact attention model & Single-head attention \\
MultiGNN & Encoder + (GCN + GAT + SAGE) + Classifier & Parallel multi-branch & Mixed pooling, concatenation \\
ResidualAttentionGNN & Input MLP + 2 GAT + Attention + Classifier & Residual + learned attention & Softmax-weighted features \\
DeepGNN & Input + 3×(GCN+GAT+SAGE) + Classifier & Hierarchical deep model & Skip connections, parallel sublayers \\
\botrule
\end{tabular}
\end{table}

\section{Results}\label{sec5}

This section presents the results of applying different machine learning approaches to the data. The main goal is to explain the performance of three groups of models: (1) classical machine learning models, (2) ensemble models, and (3) graph neural networks.

For a comprehensive assessment, the results are reported based on the standard metrics of Accuracy, Precision, Recall, and F1 score. These metrics provide both a general picture of the models’ performance and a more detailed understanding of their behavior at the class level.

The results are presented in two complementary forms: numerical tables that summarize the performance of each model, and an analytical explanation that discusses the results, highlighting strengths and weaknesses among the models.

\subsection{Results of Classical Models}\label{subsec2}

\subsubsection{Pipeline A}\label{subsubsec2}

\begin{figure}[h]
    \centering
    \includegraphics[width=0.8\textwidth]{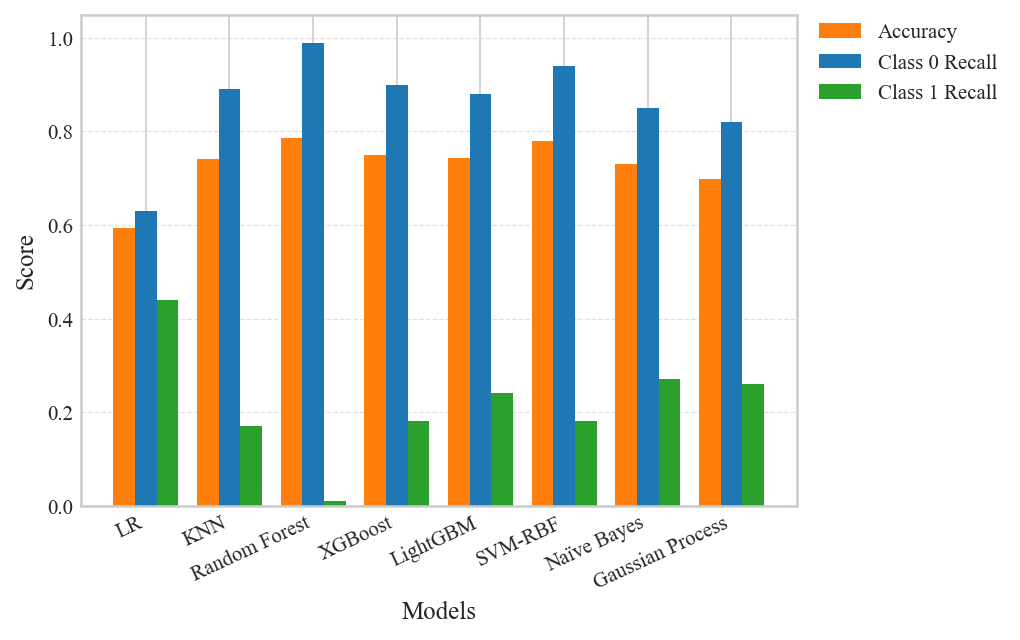}
    \caption{Comparison of classical machine learning models with the PCA-based pipeline}
    \label{fig:PCA}
\end{figure}

In the PCA-based pipeline, tree-based and kernel-based methods (such as Random Forest and SVM with RBF kernel) achieved the highest overall accuracy (the accuracy of Random Forest was reported to be around 0.79, and SVM-RBF was around 0.78). However, examination of the discriminant metrics showed that such performance was largely due to the high accuracy in identifying the major class (class 0). For example, Random Forest was able to correctly identify almost all class 0 samples (Recall $\approx 0.99$), but practically failed to identify the minority class (class 1) (Recall $\approx 0.01$). This pattern indicates a strong bias of the models towards the major class and their inefficiency in predicting purchases (minority class). Furthermore, Logistic Regression and Naïve Bayes, in contrast, showed a more balanced performance between the two classes (Class 1 Recall around 0.44), although their overall accuracy was lower. These results indicate that in unbalanced data, the accuracy measure alone cannot reflect the true power of the models in identifying important but small classes. Figure~\ref{fig:PCA} and Table~\ref{tab:pca_results} provide a summary of PCA-based pipeline results.

\begin{table}[h]
\caption{Performance of the PCA-based machine learning pipeline across different models}\label{tab:pca_results}
\centering
\renewcommand{\arraystretch}{1.2}
\begin{tabular}{@{}p{2.3cm} p{1.3cm} p{1.2cm} p{1.2cm} p{1cm} p{1.2cm} p{1.2cm} p{1cm}@{}}
\toprule
\textbf{Model} & \textbf{Accuracy} & \textbf{Class 0 Precision} & \textbf{Class 0 Recall} & \textbf{Class 0 F1} & \textbf{Class 1 Precision} & \textbf{Class 1 Recall} & \textbf{Class 1 F1} \\
\midrule
LR & 0.593 & 0.81 & 0.63 & 0.71 & 0.24 & 0.44 & 0.31 \\
KNN & 0.742 & 0.80 & 0.89 & 0.85 & 0.29 & 0.17 & 0.21 \\
Random Forest & 0.786 & 0.79 & 0.99 & 0.88 & 0.20 & 0.01 & 0.01 \\
XGBoost & 0.750 & 0.81 & 0.90 & 0.85 & 0.32 & 0.18 & 0.23 \\
LightGBM & 0.743 & 0.81 & 0.88 & 0.84 & 0.34 & 0.24 & 0.28 \\
SVM-RBF & 0.780 & 0.81 & 0.94 & 0.87 & 0.44 & 0.18 & 0.25 \\
Naïve Bayes & 0.730 & 0.82 & 0.85 & 0.83 & 0.32 & 0.27 & 0.30 \\
Gaussian Process & 0.699 & 0.81 & 0.82 & 0.81 & 0.27 & 0.26 & 0.26 \\
\bottomrule
\end{tabular}
\end{table}

\subsubsection{Pipeline B}\label{subsubsec2}

\begin{figure}[h]
    \centering
    \includegraphics[width=0.8\textwidth]{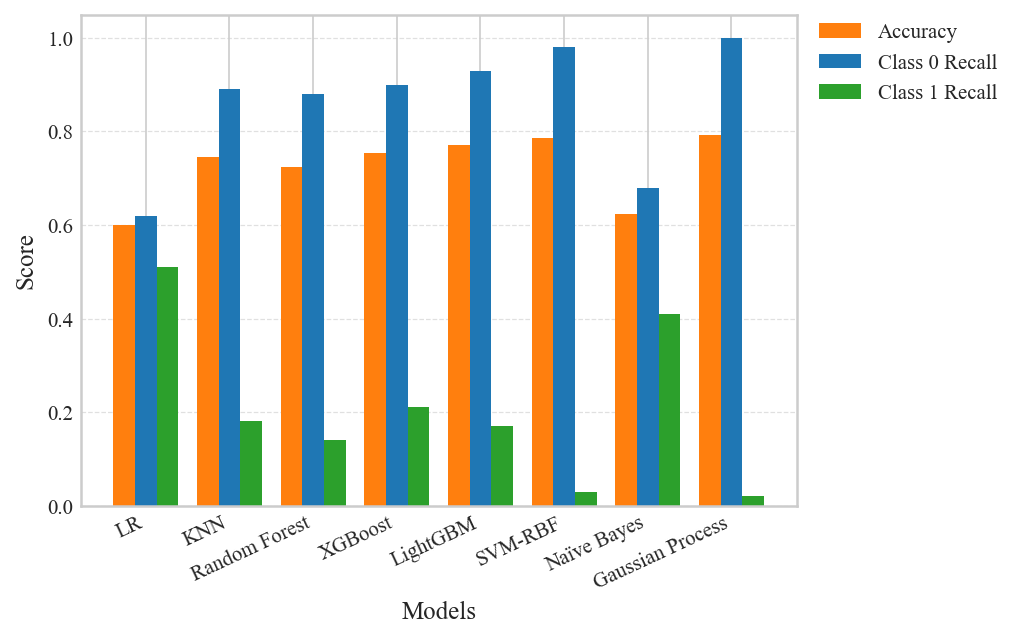}
    \caption{Comparison of classical machine learning models with the UMAP-based pipeline}
    \label{fig:UMAP}
\end{figure}

Dimensionality reduction with UMAP had different effects than PCA. In some models, such as Logistic Regression, the use of UMAP resulted in a relatively improved sensitivity to the minority class (Class 1 Recall of about 0.51 compared to 0.44 in PCA). This suggests that UMAP may have been able to reveal some of the nonlinear structures in the data that are useful for detecting purchases. However, the overall pattern remained the same; many models, such as Gaussian Process and SVM-RBF, had high overall accuracy (about 0.79) but very low sensitivity to the minority (Class 1 Recall $\approx 0.02$). Therefore, it can be concluded that although UMAP helps to improve minority detection in some cases, it alone cannot solve the problem of imbalanced data, and the choice of model and optimization criteria plays a more decisive role. In Figure~\ref{fig:UMAP} and Table~\ref{tab:umap_results} the results of the UMAP-based method can be seen.

\begin{table}[h]
\caption{Performance of the UMAP-based machine learning pipeline across different models}\label{tab:umap_results}
\centering
\renewcommand{\arraystretch}{1.2} 
\begin{tabular}{@{}p{2.3cm} p{1.3cm} p{1.2cm} p{1.2cm} p{1cm} p{1.2cm} p{1.2cm} p{1cm}@{}}
\toprule
\textbf{Model} & \textbf{Accuracy} & \textbf{Class 0 Precision} & \textbf{Class 0 Recall} & \textbf{Class 0 F1} & \textbf{Class 1 Precision} & \textbf{Class 1 Recall} & \textbf{Class 1 F1} \\
\midrule
LR & 0.599 & 0.83 & 0.62 & 0.71 & 0.26 & 0.51 & 0.35 \\
KNN & 0.745 & 0.81 & 0.89 & 0.85 & 0.31 & 0.18 & 0.23 \\
Random Forest & 0.724 & 0.79 & 0.88 & 0.83 & 0.23 & 0.14 & 0.17 \\
XGBoost & 0.753 & 0.81 & 0.90 & 0.85 & 0.35 & 0.21 & 0.26 \\
LightGBM & 0.771 & 0.81 & 0.93 & 0.87 & 0.39 & 0.17 & 0.24 \\
SVM-RBF & 0.785 & 0.79 & 0.98 & 0.88 & 0.33 & 0.03 & 0.05 \\
Naïve Bayes & 0.624 & 0.81 & 0.68 & 0.74 & 0.25 & 0.41 & 0.31 \\
Gaussian Process & 0.793 & 0.79 & 1.00 & 0.88 & 0.60 & 0.02 & 0.04 \\
\bottomrule
\end{tabular}
\end{table}

\subsubsection{Pipeline C}\label{subsubsec2}

\begin{figure}[h]
    \centering
    \includegraphics[width=0.8\textwidth]{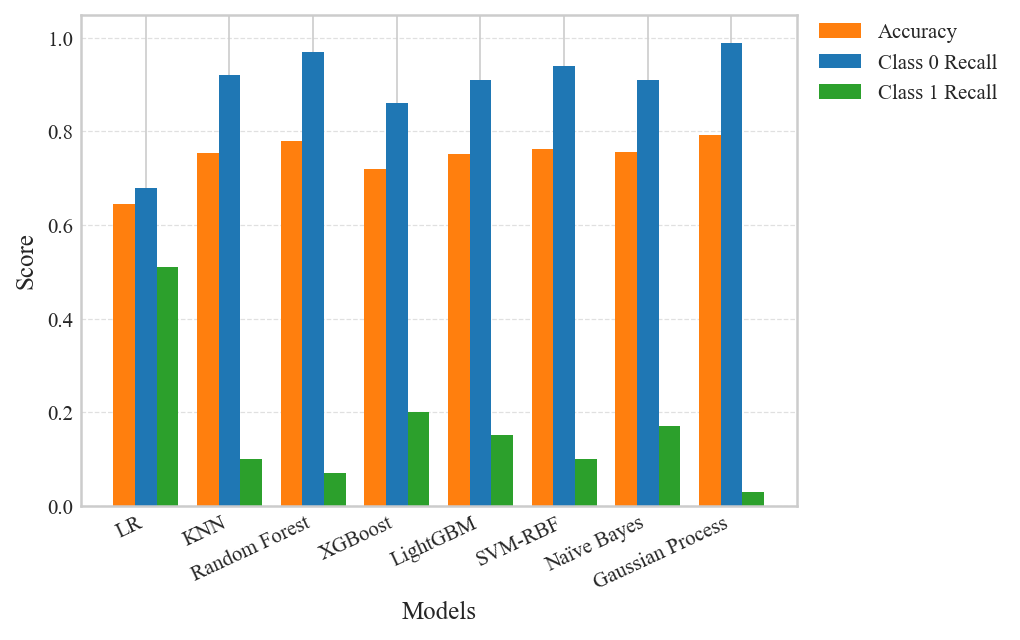}
    \caption{Comparison of classical machine learning models with the top features-based pipeline}
    \label{fig:topfeatures}
\end{figure}

The use of statistical feature selection (t-test) and then dimensionality reduction with PCA led to improvements in Logistic Regression, with a better overall accuracy and a better performance than the PCA-based and similar to the UMAP-based method on minority class detection. Logistic Regression in this case recorded a class 1 recall of 0.51, while its overall accuracy remained around 0.64. This finding suggests that noise removal and focusing on discriminant features can increase the sensitivity of the model to minorities. However, Random Forest still provided higher overall accuracy (Accuracy $\approx 0.78$), but their minority class recall remained very low. These results suggest that feature selection alone is not able to eliminate bias towards the dominant class, although it can be effective for models that are more sensitive to data structure (such as logistic regression). The summary of top features-based method is provided in Figure~\ref{fig:topfeatures} and Table~\ref{tab:top_features_results}.

\begin{table}[h]
\caption{Performance of the top features-based machine learning pipeline across different models}\label{tab:top_features_results}
\centering
\renewcommand{\arraystretch}{1.2} 
\begin{tabular}{@{}p{2.3cm} p{1.3cm} p{1.2cm} p{1.2cm} p{1cm} p{1.2cm} p{1.2cm} p{1cm}@{}}
\toprule
\textbf{Model} & \textbf{Accuracy} & \textbf{Class 0 Precision} & \textbf{Class 0 Recall} & \textbf{Class 0 F1} & \textbf{Class 1 Precision} & \textbf{Class 1 Recall} & \textbf{Class 1 F1} \\
\midrule
LR & 0.644 & 0.84 & 0.68 & 0.75 & 0.29 & 0.51 & 0.37 \\
KNN & 0.753 & 0.80 & 0.92 & 0.86 & 0.26 & 0.10 & 0.15 \\
Random Forest & 0.780 & 0.80 & 0.97 & 0.87 & 0.36 & 0.07 & 0.11 \\
XGBoost & 0.720 & 0.80 & 0.86 & 0.83 & 0.27 & 0.20 & 0.23 \\
LightGBM & 0.751 & 0.80 & 0.91 & 0.85 & 0.30 & 0.15 & 0.20 \\
SVM-RBF & 0.762 & 0.80 & 0.94 & 0.86 & 0.29 & 0.10 & 0.14 \\
Naïve Bayes & 0.757 & 0.81 & 0.91 & 0.86 & 0.34 & 0.17 & 0.23 \\
Gaussian Process & 0.793 & 0.80 & 0.99 & 0.88 & 0.57 & 0.03 & 0.06 \\
\bottomrule
\end{tabular}
\end{table}

\subsection{Results of Ensemble Models}\label{subsec2}

The results from the different pipelines show that the use of dimensionality reduction and feature selection methods has an impact on the performance of the Stacking models. As can be seen in Table 5, the overall accuracy in the three modes of PCA, UMAP, and Top Features is 0.733, 0.765, and 0.759, respectively. These results indicate that the use of UMAP in combination with Stacking provides the best overall performance and has been able to provide higher accuracy than other approaches. However, in all pipelines, it is observed that the model is mainly more successful in classifying the dominant class (class 0) and still faces weakness in recall for the minority class (class 1).

\begin{figure}[h]
    \centering
    \includegraphics[width=0.8\textwidth]{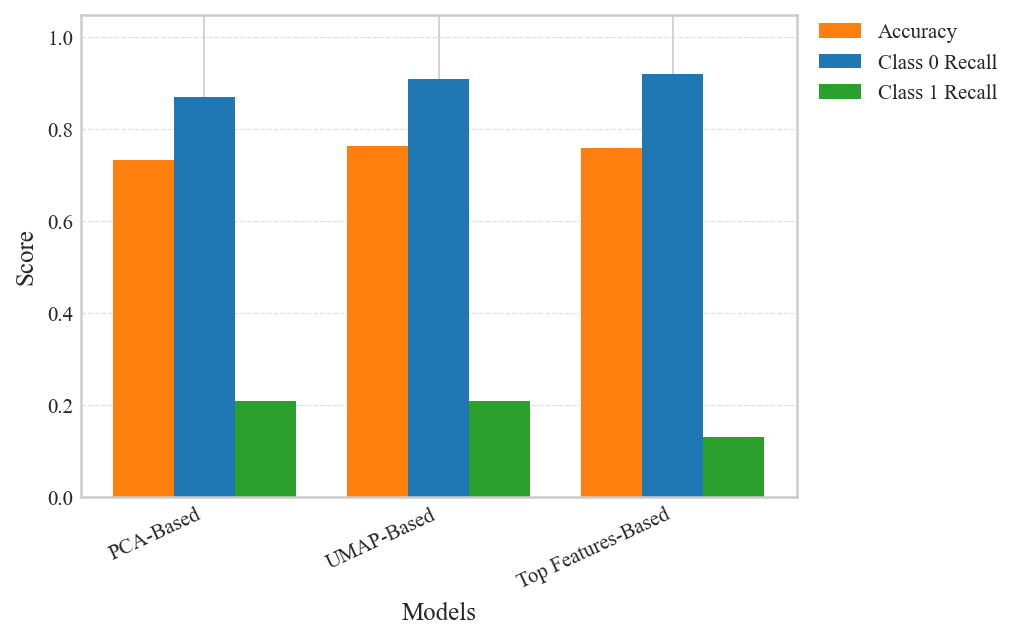}
    \caption{Comparison of ensemble models with PCA-based, UMAP-based, and top features-based pipelines}
    \label{fig:ensemble}
\end{figure}

In general, it can be concluded that the combined pipelines have been able to enhance the performance of the model by increasing the resolution of the data, especially at the level of features related to the major class. Although these approaches have had a slight effect on improving the minority class-related metrics, there is still a significant gap with a balanced classification. Therefore, the results obtained show that although the use of dimensionality reduction and feature selection can change the performance of ensemble models, complementary methods such as resampling or data imbalance-sensitive algorithms are needed to achieve optimal performance in unbalanced data. The results of the ensemble models are available in Figure~\ref{fig:ensemble} and Table~\ref{tab:ensemble_results}.

\begin{table}[h]
\caption{Performance of the ensemble machine learning models across different pipelines}\label{tab:ensemble_results}
\centering
\renewcommand{\arraystretch}{1.2} 
\begin{tabular}{@{}p{2.3cm} p{1.3cm} p{1.2cm} p{1.2cm} p{1cm} p{1.2cm} p{1.2cm} p{1cm}@{}}
\toprule
\textbf{Pipeline} & \textbf{Accuracy} & \textbf{Class 0 Precision} & \textbf{Class 0 Recall} & \textbf{Class 0 F1} & \textbf{Class 1 Precision} & \textbf{Class 1 Recall} & \textbf{Class 1 F1} \\
\midrule
A & 0.733 & 0.81 & 0.87 & 0.84 & 0.30 & 0.21 & 0.25 \\
B & 0.765 & 0.81 & 0.91 & 0.86 & 0.38 & 0.21 & 0.27 \\
C & 0.759 & 0.80 & 0.92 & 0.86 & 0.32 & 0.13 & 0.19 \\
\bottomrule
\end{tabular}
\end{table}

\subsection{Results of Graph Neural Network Models}\label{subsec2}

\begin{figure}[h]
    \centering
    \includegraphics[width=0.8\textwidth]{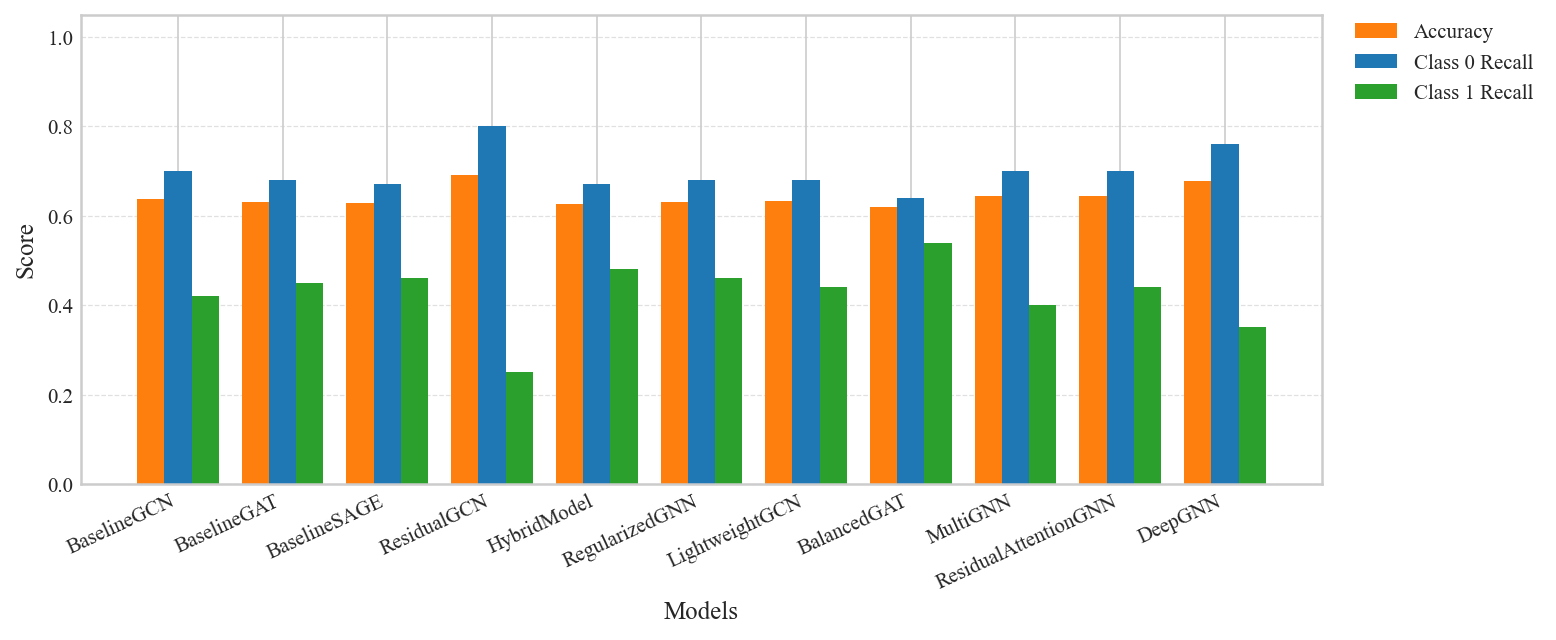}
    \caption{Comparison of GNN models with brain connectivity features}
    \label{fig:GNNs}
\end{figure}

The results obtained from the GNNs (Figure~\ref{fig:GNNs} and Table~\ref{tab:gnn_results}) show that these approaches have been able to provide acceptable performance in extracting structural features from EEG data, but still face challenges in minority class classification. As can be seen, the overall accuracy of the models is mainly in the range of 0.62 to 0.69. Among them, ResidualGCN with an accuracy of about 0.69 and DeepGNN with an accuracy of  about 0.67 have performed better than other models, indicating that increasing the network depth or using residual connections can have a positive role in improving generalizability. In contrast, baseline models such as BaselineGCN, BaselineGAT, and BaselineSAGE have obtained lower accuracy, which indicates their limitations in learning complex relationships between nodes.

In terms of class-based analysis, almost all models were successful in identifying the major class, achieving Precision and F1 values above 0.74 for this class. In contrast, the performance for the minority class was significantly weaker, with Recall for this class remaining below 0.46 in most models. Interestingly, although the BalancedGAT model had a lower overall accuracy, it provided a better Recall for the minority class, thus achieving a greater balance between the two classes. 

Overall, the results indicate that GNNs are capable of learning the association features between EEG signals, but to achieve a balanced performance across classes, complementary strategies such as data balancing methods, adaptive attention, or a combination with classical models and dimensionality reduction methods are needed. This suggests that although GNNs have been able to provide relatively good overall accuracy, the data imbalance issue remains a major challenge. 

\begin{table}[h]
\caption{Performance of the GNN machine learning models}\label{tab:gnn_results}
\centering
\renewcommand{\arraystretch}{1.2} 
\begin{tabular}{@{}p{2.3cm} p{1.3cm} p{1.2cm} p{1.2cm} p{1cm} p{1.2cm} p{1.2cm} p{1cm}@{}}
\toprule
\textbf{Model} & \textbf{Accuracy} & \textbf{Class 0 Precision} & \textbf{Class 0 Recall} & \textbf{Class 0 F1} & \textbf{Class 1 Precision} & \textbf{Class 1 Recall} & \textbf{Class 1 F1} \\
\midrule
BaselineGCN & 0.637 & 0.82 & 0.70 & 0.75 & 0.27 & 0.42 & 0.33 \\
BaselineGAT & 0.631 & 0.82 & 0.68 & 0.75 & 0.27 & 0.45 & 0.34 \\
BaselineSAGE & 0.628 & 0.83 & 0.67 & 0.74 & 0.27 & 0.46 & 0.34 \\
ResidualGCN & 0.691 & 0.80 & 0.80 & 0.81 & 0.26 & 0.25 & 0.25 \\
HybridModel & 0.627 & 0.83 & 0.67 & 0.74 & 0.27 & 0.48 & 0.35 \\
RegularizedGNN & 0.630 & 0.83 & 0.68 & 0.74 & 0.27 & 0.46 & 0.34 \\
LightweightGCN & 0.634 & 0.82 & 0.68 & 0.75 & 0.27 & 0.44 & 0.34 \\
BalancedGAT & 0.619 & 0.84 & 0.64 & 0.73 & 0.29 & 0.54 & 0.37 \\
MultiGNN & 0.644 & 0.82 & 0.70 & 0.76 & 0.27 & 0.40 & 0.32 \\
ResidualAttention\newline GNN & 0.644 & 0.83 & 0.70 & 0.76 & 0.28 & 0.44 & 0.34 \\
DeepGNN & 0.677 & 0.82 & 0.76 & 0.79 & 0.28 & 0.35 & 0.31 \\
\bottomrule
\end{tabular}
\end{table}

\subsection{Comparative Analysis of Modelling Approaches}\label{subsec2}

The results obtained show that the three main categories of approaches used, namely classical models based on EEG statistical features, ensemble models with feature selection or dimensionality reduction, and models based on graph neural networks (GNNs) using spectral features, each have their own strengths and weaknesses.
Classical models were able to provide relatively stable classification performance using statistical features extracted from EEG signals and achieved noticeable results in major class detection. These models are computationally lighter and, due to the simplicity of their structure, have relatively good generalization capabilities in limited data conditions, but they have limitations in learning complex and nonlinear patterns between EEG channels.

Ensemble models can show a balance in terms of accuracy and minority class prediction, while they are using different dimensionality reduction methods, such as PCA and UMAP, proposed in the pipelines. The stacking model, which used UMAP as a dimensionality reduction method, achieved higher accuracy than other ensemble models that were trained in the different pipelines. However, like classical models, these models still faced challenges in identifying the minority class.

Unlike the classical and ensemble models, GNN-based models, which used EEG spectral features and constructed brain connectivity graphs for training, were able to provide deeper insights in terms of the interactions between the nodes. However, their overall accuracy remained lower than the ensemble models in most cases, and only some more advanced architectures (such as ResidualGCN and DeepGNN) were able to provide more competitive results. Overall, the minority class prediction results in graph-based models are more remarkable than other methods, and in cases such as BalancedGAT, it was able to obtain higher Recall for the minority class.

Overall, classical and ensemble models based on EEG statistical features still succeeded in providing better accuracy, while GNN models, despite lower accuracy, have higher potential in learning complex relationships between channels and exploiting spectral structures, and providing a better performance on minor class prediction. The results suggest that in future research combination of classical approaches with graph-based architectures can be utilized to develop hybrid methods, which can take the advantages of both methods and improve the predictions for the minority class.
\section{Discussion}\label{sec7}

The obtained results from the utilized models show that classical, ensemble, and GNN models each have their own abilities in analysing and modelling EEG signals collected by marketing stimuli. Overall, classical models show a better accuracy, while GNNs show a better performance in identifying the minority class by modelling the structural relationships between brain channels.
This research shows that GNN models performed better in identifying complex cognitive patterns, furthermore, their performance on the minority class is better than the classical and ensemble models that combined with different feature selection models.

\subsection{Theoretical Implications}\label{subsec2}

From a theoretical perspective, this study shows that graph-based modelling can be introduced as a novel method in neuromarketing tasks. Instead of focusing on time domain and frequency domain features, using brain connectivity features can provide a deeper insight into the consumer decision-making process. This can be utilized in the development of data-driven cognitive models in marketing and neuroscience.

GNNs are able to model the topological connections between electrodes that collect the EEG data from the brain. Due to this characteristic of GNNs, they can model the interactions between different nodes of the graph, which are the electrodes in an EEG-based task. These research findings introduce a new idea in neuromarketing and give a direction to future neuromarketing studies.

Overall, by introducing a graph-based framework for EEG analysis in neuromarketing, this study provides a new approach for modelling consumer brain data collected by using marketing stimuli. These findings can lead to developing novel theoretical models that can explain the consumer decision-making process as a brain connectivity network.

\subsection{Practical Implications}\label{subsec2}

From a practical perspective, the results of this study can provide a deeper insight for companies, brands, and marketing specialists. The proposed models have the ability to classify consumers' brain EEG signals in response to a marketing stimulus. This can lead to new developments for optimizing marketing campaigns and product design.

In addition, using machine learning models could be helpful for companies to segment their consumers into different groups by using brain responses. Thus, personalized marketing can be implemented on a deeper level. Furthermore, companies can use EEG-based methods to assess the effectiveness of pre-launch advertising by showing multiple versions of an advertisement to different groups and using EEG to select the most effective version.

\section{Conclusion}\label{sec7}

In this study, different machine learning methods and pipelines have been utilized for predicting consumer behavior by using an EEG-based dataset with neuromarketing stimuli. Two main methods have been implemented and compared: (1) classical and ensemble models with three different pipelines that used statistical EEG features, and (2) graph-based models that used spectral features with brain connectivity representations and different graph neural network architectures. The results showed that, while the classical models and pipeline have a better performance in terms of accuracy, the GNNs perform better in predicting the minor class.

The findings of this study introduced the remarkable potential of graph neural networks in the EEG-based tasks by modelling the brain connectivity, which shows the interactions between the nodes. Due to the computational simplicity and interpretability, classical models can remain competitive in tasks where these measures are important, and this shows that the method selection can depend on the needs of a task and the problem specifications.

In spite of the significant results and methods that have been used in this study, the limitations should be mentioned. The study was conducted by using a limited dataset, and due to this, the results may affect the model in terms of robustness and generalizability. Furthermore, in future studies, different connectivity features can be examined to develop richer models and more accurate predictions.

Overall, this study provides a deeper insight into the effectiveness of graph-based approaches in neuromarketing applications, compares classical, ensemble, and GNN models from a neuromarketing perspective, and shows the applications of machine learning and deep learning methods in EEG-based consumer behavior predictions. The future research can be pursued by combining classic and deep learning models, using a wide range of datasets, and developing more robust models, utilizing different methods for construction on the brain connectivity graphs, and integrating EEG with other modalities such as eye tracking, physiological data, fMRI, or fNIRS data. These approaches can cause advancements in the field of neuromarketing in both practical insights and scientific prospective.

\textbf{}

{[}1{]} P. Singh, I. Alhassan, and L. Khoshaim, ``What Do You Need to
Know? A Systematic Review and Research Agenda on Neuromarketing
Discipline,'' Dec. 01, 2023, \emph{Multidisciplinary Digital Publishing
Institute (MDPI)}. doi: 10.3390/jtaer18040101.

{[}2{]} L. Jing, C. Tian, S. He, D. Feng, S. Jiang, and C. Lu,
``Data-driven implicit design preference prediction model for product
concept evaluation via BP neural network and EEG,'' \emph{Advanced
Engineering Informatics}, vol. 58, p. 102213, Oct. 2023, doi:
10.1016/j.aei.2023.102213.

{[}3{]} Z. Zhu, Y. Jin, Y. Su, K. Jia, C. L. Lin, and X. Liu,
``Bibliometric-Based Evaluation of the Neuromarketing Research Trend:
2010--2021,'' \emph{Front Psychol}, vol. 13, p. 872468, Aug. 2022, doi:
10.3389/fpsyg.2022.872468.

{[}4{]} A. A. Mansor, S. M. Isa, and S. S. M. Noor, ``P300 and
decision-making in neuromarketing,'' \emph{Neuroscience Research Notes},
vol. 4, no. 3, pp. 21--26, Sep. 2021, doi: 10.31117/NEUROSCIRN.V4I3.83.

{[}5{]} G. Pei and T. Li, ``A Literature Review of EEG-Based Affective
Computing in Marketing,'' \emph{Front Psychol}, vol. 12, p. 602843, Mar.
2021, doi: 10.3389/fpsyg.2021.602843.

{[}6{]} J. P. M. dos Santos, H. Ferreira, J. Reis, D. Prata, S. P.
Simões, and I. D. Borges, ``The Use of Consumer Neuroscience Knowledge
in Improving Real Promotional Media: The Case of Worten,'' 2020, pp.
202--218. doi: 10.1007/978-981-15-1564-4\_20.

{[}7{]} A. H. Alsharif, N. Z. Md Salleh, R. Baharun, and A. Rami Hashem
E, ``Neuromarketing research in the last five years: a bibliometric
analysis,'' \emph{Cogent Business \& Management}, vol. 8, no. 1, Jan.
2021, doi: 10.1080/23311975.2021.1978620.

{[}8{]} V. Khurana \emph{et al.}, ``A Survey on Neuromarketing Using EEG
Signals,'' \emph{IEEE Trans Cogn Dev Syst}, vol. 13, no. 4, pp.
732--749, Dec. 2021, doi: 10.1109/TCDS.2021.3065200.

{[}9{]} M. F. K. Khondakar \emph{et al.}, ``A systematic review on
EEG-based neuromarketing: recent trends and analyzing techniques,'' Dec.
01, 2024, \emph{Springer Science and Business Media Deutschland GmbH}.
doi: 10.1186/s40708-024-00229-8.

{[}10{]} A. Ali \emph{et al.}, ``EEG Signals Based Choice Classification
for Neuromarketing Applications,'' 2022, pp. 371--394. doi:
10.1007/978-3-030-76653-5\_20.

{[}11{]} M. Esteban-Bravo and J. M. Vidal-Sanz, \emph{Marketing research
methods: quantitative and qualitative approaches}. Cambridge University
Press, 2021.

{[}12{]} T. Zhao \emph{et al.}, ``AutoML-based EEG signal analysis in
neuro-marketing classification using biclustering method,'' \emph{Appl
Soft Comput}, vol. 181, p. 113412, Sep. 2025, doi:
10.1016/j.asoc.2025.113412.

{[}13{]} S. Khaneja and T. Arora, ``The potential of neuroscience in
transforming business: a meta-analysis,'' \emph{Future Business
Journal}, vol. 10, no. 1, p. 77, Dec. 2024, doi:
10.1186/s43093-024-00369-7.

{[}14{]} A. Stasi \emph{et al.}, ``Neuromarketing empirical approaches
and food choice: A systematic review,'' \emph{Food Research
International}, vol. 108, pp. 650--664, Jun. 2018, doi:
10.1016/j.foodres.2017.11.049.

{[}15{]} C. M. L. Cruz, J. F. De Medeiros, L. C. R. Hermes, A. Marcon,
and É. Marcon, ``Neuromarketing and the advances in the consumer
behaviour studies: a systematic review of the literature,''
\emph{International Journal of Business and Globalisation}, vol. 17, no.
3, p. 330, 2016, doi: 10.1504/IJBG.2016.078842.

{[}16{]} V. Duarte, S. Zuniga-Jara, and S. Contreras, ``Machine Learning
and Marketing: A Systematic Literature Review,'' \emph{IEEE Access},
vol. 10, pp. 93273--93288, 2022, doi: 10.1109/ACCESS.2022.3202896.

{[}17{]} P. Tirandazi, S. M. H. Bamakan, and A. Toghroljerdi, ``A review
of studies on internet of everything as an enabler of neuromarketing
methods and techniques,'' \emph{Journal of Supercomputing}, vol. 79, no.
7, pp. 7835--7876, May 2023, doi: 10.1007/s11227-022-04988-1.

{[}18{]} D. Juarez, V. Tur-Viñes, and A. Mengual, ``Neuromarketing
Applied to Educational Toy Packaging,'' \emph{Front Psychol}, vol. 11,
Aug. 2020, doi: 10.3389/fpsyg.2020.02077.

{[}19{]} S. Zhu, J. Qi, J. Hu, and S. Hao, ``A new approach for product
evaluation based on integration of EEG and eye-tracking,''
\emph{Advanced Engineering Informatics}, vol. 52, p. 101601, Apr. 2022,
doi: 10.1016/j.aei.2022.101601.

{[}20{]} M. Peng, Z. Xu, and H. Huang, ``How Does Information Overload
Affect Consumers' Online Decision Process? An Event-Related Potentials
Study,'' \emph{Front Neurosci}, vol. 15, Oct. 2021, doi:
10.3389/fnins.2021.695852.

{[}21{]} X. Li, D. B. Luh, and Z. Chen, ``A Systematic Review and
Meta-Analysis of Eye-Tracking Studies for Consumers' Visual Attention in
Online Shopping,'' \emph{Information Technology and Control}, vol. 53,
no. 1, pp. 187--205, Aug. 2024, doi: 10.5755/j01.itc.53.1.34855.

{[}22{]} S. Lou, Y. Feng, H. Zheng, Y. Gao, and J. Tan, ``Data-driven
customer requirements discernment in the product lifecycle management
via intuitionistic fuzzy sets and electroencephalogram,'' \emph{J Intell
Manuf}, vol. 31, no. 7, pp. 1721--1736, Oct. 2020, doi:
10.1007/s10845-018-1395-x.

{[}23{]} J. P. Marques dos Santos and J. D. Marques dos Santos,
``Explainable artificial intelligence (xAI) in neuromarketing/consumer
neuroscience: an fMRI study on brand perception,'' \emph{Front Hum
Neurosci}, vol. 18, p. 1305164, Mar. 2024, doi:
10.3389/fnhum.2024.1305164.

{[}24{]} E. L. E. De Vries, B. M. Fennis, T. H. A. Bijmolt, G. J. Ter
Horst, and J.-B. C. Marsman, ``Friends with benefits: Behavioral and
fMRI studies on the effect of friendship reminders on self-control for
compulsive and non-compulsive buyers,'' \emph{International Journal of
Research in Marketing}, vol. 35, no. 2, pp. 336--358, Jun. 2018, doi:
10.1016/j.ijresmar.2017.12.004.

{[}25{]} S. J. Bak, J. Shin, D. Kim, and S. Kim, ``Analysis of consumer
purchase intentions using functional near-infrared spectroscopy(fNIRS):
A neuromarketing study on the aesthetic packaging of Korean red ginseng
products,'' \emph{PLoS One}, vol. 20, no. 6 June, p. e0326213, Jun.
2025, doi: 10.1371/journal.pone.0326213.

{[}26{]} A. Gorin \emph{et al.}, ``Neural correlates of the non-optimal
price: an MEG/EEG study,'' \emph{Front Hum Neurosci}, vol. 19, p.
1470662, Jan. 2025, doi: 10.3389/fnhum.2025.1470662.

{[}27{]} A. H. Alsharif and S. Mohd Isa, ``Revolutionizing consumer
insights: the impact of fMRI in neuromarketing research,'' \emph{Future
Business Journal}, vol. 10, no. 1, p. 79, Dec. 2024, doi:
10.1186/s43093-024-00371-z.

{[}28{]} A. Hassani, A. Hekmatmanesh, and A. M. Nasrabadi,
``Discrimination of Customers Decision-Making in a Like/Dislike Shopping
Activity Based on Genders: A Neuromarketing Study,'' \emph{IEEE Access},
vol. 10, pp. 92454--92466, 2022, doi: 10.1109/ACCESS.2022.3201488.

{[}29{]} S. Pal, P. Das, R. Sahu, and S. R. Dash, ``Study of
Neuromarketing With EEG Signals and Machine Learning Techniques,'' in
\emph{Machine Learning for Healthcare Applications}, wiley, 2021, pp.
33--56. doi: 10.1002/9781119792611.ch3.

{[}30{]} F. P. Kalaganis, K. Georgiadis, V. P. Oikonomou, N. A.
Laskaris, S. Nikolopoulos, and I. Kompatsiaris, ``Unlocking the
Subconscious Consumer Bias: A Survey on the Past, Present, and Future of
Hybrid EEG Schemes in Neuromarketing,'' \emph{Frontiers in
Neuroergonomics}, vol. 2, May 2021, doi: 10.3389/fnrgo.2021.672982.

{[}31{]} L. Robaina-Calderín and J. D. Martín-Santana, ``A review of
research on neuromarketing using content analysis: key approaches and
new avenues,'' \emph{Cogn Neurodyn}, vol. 15, no. 6, pp. 923--938, Dec.
2021, doi: 10.1007/s11571-021-09693-y.

{[}32{]} M. Saeidi \emph{et al.}, ``Neural decoding of eeg signals with
machine learning: A systematic review,'' Nov. 01, 2021, \emph{MDPI}.
doi: 10.3390/brainsci11111525.

{[}33{]} A. Bazzani, S. Ravaioli, L. Trieste, U. Faraguna, and G.
Turchetti, ``Is EEG Suitable for Marketing Research? A Systematic
Review,'' \emph{Front Neurosci}, vol. 14, Dec. 2020, doi:
10.3389/fnins.2020.594566.

{[}34{]} L. Casado‐Aranda, J. Sánchez‐Fernández, E. Bigne, and A.
Smidts, ``The application of neuromarketing tools in communication
research: A comprehensive review of trends,'' \emph{Psychol Mark}, vol.
40, no. 9, pp. 1737--1756, Sep. 2023, doi: 10.1002/mar.21832.

{[}35{]} A. H. Alsharif \emph{et al.}, ``Neuroimaging Techniques in
Advertising Research: Main Applications, Development, and Brain Regions
and Processes,'' \emph{Sustainability}, vol. 13, no. 11, p. 6488, Jun.
2021, doi: 10.3390/su13116488.

{[}36{]} M. Alimardani and M. Kaba, ``Deep Learning for Neuromarketing;
Classification of User Preference using EEG Signals,'' in \emph{12th
Augmented Human International Conference}, New York, NY, USA: ACM, May
2021, pp. 1--7. doi: 10.1145/3460881.3460930.

{[}37{]} N. K. Murungi, M. V. Pham, X. C. Dai, X. Qu, and C. Dai,
``Empowering Computer Science Students in Electroencephalography (EEG)
Analysis: A Review of Machine Learning Algorithms for EEG Datasets,'' in
\emph{Proceedings of KDD Undergraduate Consortium (KDDUC '23).}, 2023.
doi: XXXXXXX.XXXXXXX.

{[}38{]} S. M. Atoar Rahman \emph{et al.}, ``Advancement in Graph Neural
Networks for EEG Signal Analysis and Application: A Review,'' \emph{IEEE
Access}, vol. 13, pp. 50167--50187, 2025, doi:
10.1109/ACCESS.2025.3549120.

{[}39{]} H. Xiong, Y. Yan, Y. Chen, and J. Liu, ``Graph convolution
network-based eeg signal analysis: a review,'' \emph{Med Biol Eng
Comput}, vol. 63, no. 6, pp. 1609--1625, Jun. 2025, doi:
10.1007/s11517-025-03295-0.

{[}40{]} H. Mohammadi and W. Karwowski, ``Graph Neural Networks in Brain
Connectivity Studies: Methods, Challenges, and Future Directions,''
\emph{Brain Sci}, vol. 15, no. 1, p. 17, Dec. 2024, doi:
10.3390/brainsci15010017.

{[}41{]} L. E. Ismail and W. Karwowski, ``A Graph Theory-Based Modeling
of Functional Brain Connectivity Based on EEG: A Systematic Review in
the Context of Neuroergonomics,'' \emph{IEEE Access}, vol. 8, pp.
155103--155135, 2020, doi: 10.1109/ACCESS.2020.3018995.

{[}42{]} D. Klepl, M. Wu, and F. He, ``Graph Neural Network-Based EEG
Classification: A Survey,'' \emph{IEEE Transactions on Neural Systems
and Rehabilitation Engineering}, vol. 32, pp. 493--503, 2024, doi:
10.1109/TNSRE.2024.3355750.

{[}43{]} L. Rui, Wang Yiting, and Lu Bao-Liang, ``A Multi-Domain
Adaptive Graph Convolutional Network for EEG-based Emotion
Recognition,'' in \emph{Proceedings of the 29th ACM International
Conference on Multimedia}, 2021, pp. 5565--5573.

{[}44{]} Y. Yin, X. Zheng, B. Hu, Y. Zhang, and X. Cui, ``EEG emotion
recognition using fusion model of graph convolutional neural networks
and LSTM,'' \emph{Appl Soft Comput}, vol. 100, p. 106954, Mar. 2021,
doi: 10.1016/j.asoc.2020.106954.

{[}45{]} D. Pan \emph{et al.}, ``MSFR-GCN: A Multi-Scale Feature
Reconstruction Graph Convolutional Network for EEG Emotion and Cognition
Recognition,'' \emph{IEEE Transactions on Neural Systems and
Rehabilitation Engineering}, vol. 31, pp. 3245--3254, 2023, doi:
10.1109/TNSRE.2023.3304660.

{[}46{]} S. Tang \emph{et al.}, ``Self-Supervised Graph Neural Networks
for Improved Electroencephalographic Seizure Analysis,'' in
\emph{International Conference on Learning Representations}, 2022.

{[}47{]} D. Klepl, F. He, M. Wu, D. J. Blackburn, and P. Sarrigiannis,
``EEG-Based Graph Neural Network Classification of Alzheimer's Disease:
An Empirical Evaluation of Functional Connectivity Methods,'' \emph{IEEE
Transactions on Neural Systems and Rehabilitation Engineering}, vol. 30,
pp. 2651--2660, 2022, doi: 10.1109/TNSRE.2022.3204913.

{[}48{]} A. Azimi and M. P. Afshar, ``Machine Learning Techniques for
Eeg-Based Neuromarketing: A Systematic Literature Review,'' in
\emph{International Conference on Web Research, ICWR}, Institute of
Electrical and Electronics Engineers Inc., Apr. 2025, pp. 175--182. doi:
10.1109/ICWR65219.2025.11006237.

{[}49{]} K. Georgiadis \emph{et al.}, ``NeuMa - the absolute
Neuromarketing dataset en route to an holistic understanding of consumer
behaviour,'' \emph{Sci Data}, vol. 10, no. 1, p. 508, Aug. 2023, doi:
10.1038/s41597-023-02392-9.

{[}50{]} M. Yadava, P. Kumar, R. Saini, P. P. Roy, and D. Prosad Dogra,
``Analysis of EEG signals and its application to neuromarketing,''
\emph{Multimed Tools Appl}, vol. 76, no. 18, pp. 19087--19111, Sep.
2017, doi: 10.1007/s11042-017-4580-6.

{[}51{]} A. Özbeyaz, ``EEG-Based classification of branded and unbranded
stimuli associating with smartphone products: comparison of several
machine learning algorithms,'' \emph{Neural Comput Appl}, vol. 33, no.
9, pp. 4579--4593, May 2021, doi: 10.1007/s00521-021-05779-0.

{[}52{]} L. Zeng, M. Lin, K. Xiao, J. Wang, and H. Zhou, ``Like/Dislike
Prediction for Sport Shoes With Electroencephalography: An Application
of Neuromarketing,'' \emph{Front Hum Neurosci}, vol. 15, p. 793952, Jan.
2022, doi: 10.3389/fnhum.2021.793952.

{[}53{]} Z. Xu and S. Liu, ``Decoding consumer purchase decisions:
exploring the predictive power of EEG features in online shopping
environments using machine learning,'' \emph{Humanit Soc Sci Commun},
vol. 11, no. 1, pp. 1--13, Sep. 2024, doi: 10.1057/s41599-024-03691-1.

{[}54{]} F. R. Mashrur \emph{et al.}, ``BCI-Based Consumers' Choice
Prediction From EEG Signals: An Intelligent Neuromarketing Framework,''
\emph{Front Hum Neurosci}, vol. 16, p. 861270, May 2022, doi:
10.3389/fnhum.2022.861270.

{[}55{]} M. Maram, M. A. Khalil, and K. George, ``Analysis of Consumer
Coffee Brand Preferences Using Brain-Computer Interface and Deep
Learning,'' in \emph{2023 IEEE 7th International Conference on
Information Technology, Information Systems and Electrical Engineering
(ICITISEE)}, Purwokerto: IEEE, 2023, pp. 227--232. doi:
10.1109/ICITISEE58992.2023.10404368.

{[}56{]} J. J. González-España, K. J. Back, D. Reynolds, and J. L.
Contreras-Vidal, ``Decoding Taste from EEG: Gustatory Evoked Potentials
During Wine Tasting,'' in \emph{2023 IEEE International Conference on
Systems, Man, and Cybernetics (SMC)}, Honolulu: IEEE, 2023, pp.
4253--4258. doi: 10.1109/SMC53992.2023.10394408.

{[}57{]} Y. Guo, X. Xia, Y. Shi, Y. Ying, and H. Men, ``Olfactory EEG
induced by odor: Used for food identification and pleasure analysis,''
\emph{Food Chem}, vol. 455, p. 139816, Oct. 2024, doi:
10.1016/j.foodchem.2024.139816.

{[}58{]} S. P. Akbugday, B. Akbugday, F. Yeganli, A. Akan, and R.
Sadikzade, ``Decoding Olfactory EEG Signals Using Multi-Domain Features
and Machine Learning,'' in \emph{2024 Medical Technologies Congress
(TIPTEKNO)}, Mugla: IEEE, 2024, pp. 1--4. doi:
10.1109/TIPTEKNO63488.2024.10755366.

{[}59{]} B. Akbugday, S. P. Akbugday, R. Sadikzade, A. Akan, and S.
Unal, ``Detection of Olfactory Stimulus in Electroencephalogram Signals
Using Machine and Deep Learning Methods,'' \emph{Electrica}, vol. 24,
no. 1, pp. 175--182, Jan. 2024, doi: 10.5152/electrica.2024.23111.

{[}60{]} H. Vedder, F. Stano, and M. Knierim, ``Classification of Music
Preferences Using EEG Data in Machine Learning Models,'' in \emph{Mensch
und Computer 2024-Workshopband}, Gesellschaft für Informatik e.V., 2024.
doi: 10.18420/muc2024-mci-src-324.

{[}61{]} Y. P. Lin, Y. H. Yang, and T. P. Jung, ``Fusion of
electroencephalographic dynamics and musical contents for estimating
emotional responses in music listening,'' \emph{Front Neurosci}, vol. 8,
p. 94, 2014, doi: 10.3389/fnins.2014.00094.

{[}62{]} M. F. K. Khondakar, T. G. Trovee, M. Hasan, M. H. Sarowar, M.
H. Chowdhury, and Q. D. Hossain, ``A Comparative Analysis of Different
Pre-Processing Pipelines for EEG-Based Preference Prediction in
Neuromarketing,'' in \emph{2023 IEEE Pune Section International
Conference (PuneCon)}, Pune: IEEE, 2023, pp. 1-7,. doi:
10.1109/PuneCon58714.2023.10450086.

{[}63{]} A. Hakim, I. Golan, S. Yefet, and D. J. Levy, ``DeePay: deep
learning decodes EEG to predict consumer's willingness to pay for
neuromarketing,'' \emph{Front Hum Neurosci}, vol. 17, p. 1153413, 2023,
doi: 10.3389/fnhum.2023.1153413.

{[}64{]} J. Guixeres \emph{et al.}, ``Consumer neuroscience-based
metrics predict recall, liking and viewing rates in online
advertising,'' \emph{Front Psychol}, vol. 8, p. 1808, Oct. 2017, doi:
10.3389/fpsyg.2017.01808.

{[}65{]} O. Bălan, G. Moise, L. Petrescu, A. Moldoveanu, M. Leordeanu,
and F. Moldoveanu, ``Emotion classification based on biophysical signals
and machine learning techniques,'' \emph{Symmetry (Basel)}, vol. 12, no.
1, p. 21, 2020, doi: 10.3390/sym12010021.

{[}66{]} F. R. Mashrur \emph{et al.}, ``Intelligent neuromarketing
framework for consumers' preference prediction from
electroencephalography signals and eye tracking,'' \emph{Journal of
Consumer Behaviour}, vol. 23, no. 3, pp. 1146--1157, May 2024, doi:
10.1002/cb.2253.

{[}67{]} K. Georgiadis, F. P. Kalaganis, V. P. Oikonomou, S.
Nikolopoulos, N. A. Laskaris, and I. Kompatsiaris, ``NeuMark: A
Riemannian EEG Analysis Framework for Neuromarketing,'' \emph{Brain
Inform}, vol. 9, no. 1, p. 22, Dec. 2022, doi:
10.1186/s40708-022-00171-7.

{[}68{]} Z. Wei, C. Wu, X. Wang, A. Supratak, P. Wang, and Y. Guo,
``Using support vector machine on EEG for advertisement impact
assessment,'' \emph{Front Neurosci}, vol. 12, p. 76, Mar. 2018, doi:
10.3389/fnins.2018.00076.

{[}69{]} M. Aldayel, M. Ykhlef, and A. Al-Nafjan, ``Consumers'
Preference Recognition Based on Brain--Computer Interfaces: Advances,
Trends, and Applications,'' \emph{Arab J Sci Eng}, vol. 46, no. 9, pp.
8983--8997, Sep. 2021, doi: 10.1007/s13369-021-05695-4.

{[}70{]} M. Aldayel, M. Ykhlef, and A. Al-Nafjan, ``Deep learning for
EEG-based preference classification in neuromarketing,'' \emph{Applied
Sciences (Switzerland)}, vol. 10, no. 4, Feb. 2020, doi:
10.3390/app10041525.

{[}71{]} H. Göker, ``Multi-channel EEG-based classification of consumer
preferences using multitaper spectral analysis and deep learning
model,'' \emph{Multimed Tools Appl}, vol. 83, no. 14, pp. 40753--40771,
Apr. 2024, doi: 10.1007/s11042-023-17114-x.

{[}72{]} M. Aldayel, M. Ykhlef, and A. Al-Nafjan, ``Recognition of
Consumer Preference by Analysis and Classification EEG Signals,''
\emph{Front Hum Neurosci}, vol. 14, p. 604639, Jan. 2021, doi:
10.3389/fnhum.2020.604639.

{[}73{]} S. M. A. Shah \emph{et al.}, ``An Ensemble Model for Consumer
Emotion Prediction Using EEG Signals for Neuromarketing Applications,''
\emph{Sensors}, vol. 22, no. 24, p. 9744, Dec. 2022, doi:
10.3390/s22249744.

{[}74{]} A. Ullah \emph{et al.}, ``Neuromarketing Solutions based on EEG
Signal Analysis using Machine Learning,'' \emph{Int J Adv Comput Sci
Appl}, vol. 13, no. 1, p. 2022, 2022, {[}Online{]}. Available:
www.ijacsa.thesai.org

{[}75{]} H. Azadravesh, R. Sheibani, and Y. Forghani, ``Predicted
consumer buying behavior in neural marketing based on convolutional
neural network and short-term long-term memory,'' \emph{Multimed Tools
Appl}, pp. 1--17, 2024, doi: 10.1007/s11042-024-19742-3.

{[}76{]} D. Panda, D. Das Chakladar, and T. Dasgupta, ``Prediction of
consumer preference for the bottom of the pyramid using EEG-based deep
model,'' \emph{International Journal of Computational Science and
Engineering}, vol. 24, no. 5, pp. 439--449, 2021.

{[}77{]} G. Chiarion, L. Sparacino, Y. Antonacci, L. Faes, and L. Mesin,
``Connectivity Analysis in EEG Data: A Tutorial Review of the State of
the Art and Emerging Trends,'' \emph{Bioengineering}, vol. 10, no. 3, p.
372, Mar. 2023, doi: 10.3390/bioengineering10030372.

{[}78{]} A. Hakim, S. Klorfeld, T. Sela, D. Friedman, M. Shabat-Simon,
and D. J. Levy, ``Machines learn neuromarketing: Improving preference
prediction from self-reports using multiple EEG measures and machine
learning,'' \emph{International Journal of Research in Marketing}, vol.
38, no. 3, pp. 770--791, Sep. 2021, doi: 10.1016/j.ijresmar.2020.10.005.

{[}79{]} J. Teo, L. Hou, and J. Mountstephens, ``Preference
Classification Using Electroencephalography (EEG) and Deep Learning,''
\emph{Journal of Telecommunication, Electronic and Computer Engineering
(JTEC)}, vol. 10, no. 1--11, pp. 87--91, Feb. 2018.

{[}80{]} T. Chen, Z. Lin, K. Ren, and T. Ren, ``Neuromarketing research
and EEG signal analysis,'' in \emph{International Conference on Computer
Graphics, Artificial Intelligence, and Data Processing (ICCAID 2022)},
SPIE, May 2023, pp. 959--967. doi: 10.1117/12.2674634.

{[}81{]} S. M. Usman, S. M. Ali Shah, O. C. Edo, and J. Emakhu, ``A Deep
Learning Model for Classification of EEG Signals for Neuromarketing,''
in \emph{2023 International Conference on IT Innovation and Knowledge
Discovery (ITIKD)}, Manama: IEEE, 2023, pp. 1--6. doi:
10.1109/ITIKD56332.2023.10100014.

{[}82{]} G. Ruffini, D. Ibañez, M. Castellano, S. Dunne, and A.
Soria-Frisch, ``EEG-driven RNN Classification for Prognosis of
Neurodegeneration in At-Risk Patients,'' in \emph{Artificial Neural
Networks and Machine Learning -- ICANN 2016}, A. E. P. Villa, P.
Masulli, and A. J. Pons Rivero, Eds., Cham: Springer International
Publishing, 2016, pp. 306--313.

{[}83{]} K. R. Hole and N. Agnihotri, ``EEG-Based Brain Disease
Classification: A Proposed Model Using CNN-RNN Ensemble Process,'' in
\emph{2024 11th International Conference on Reliability, Infocom
Technologies and Optimization (Trends and Future Directions), ICRITO
2024}, IEEE, Mar. 2024, pp. 1--6. doi:
10.1109/ICRITO61523.2024.10522464.

{[}84{]} A. Demir, T. Koike-Akino, Y. Wang, and D. Erdogmus, ``EEG-GAT:
Graph Attention Networks for Classification of Electroencephalogram
(EEG) Signals,'' in \emph{2022 44th Annual International Conference of
the IEEE Engineering in Medicine \& Biology Society (EMBC)}, IEEE, Jul.
2022, pp. 30--35. doi: 10.1109/EMBC48229.2022.9871984.

{[}85{]} M. Geravanchizadeh, A. Shaygan Asl, and S. Danishvar,
``Selective Auditory Attention Detection Using Combined Transformer and
Convolutional Graph Neural Networks,'' \emph{Bioengineering}, vol. 11,
no. 12, p. 1216, Nov. 2024, doi: 10.3390/bioengineering11121216.

{[}86{]} S. Blum, N. S. J. Jacobsen, M. G. Bleichner, and S. Debener,
``A Riemannian Modification of Artifact Subspace Reconstruction for EEG
Artifact Handling,'' \emph{Front Hum Neurosci}, vol. 13, Apr. 2019, doi:
10.3389/fnhum.2019.00141.

{[}87{]} I. Daly, R. Scherer, M. Billinger, and G. Muller-Putz, ``FORCe:
Fully Online and Automated Artifact Removal for Brain-Computer
Interfacing,'' \emph{IEEE Transactions on Neural Systems and
Rehabilitation Engineering}, vol. 23, no. 5, pp. 725--736, Sep. 2015,
doi: 10.1109/TNSRE.2014.2346621.

~

\end{document}